\documentclass[10pt,twocolumn,letterpaper]{article}

\usepackage{cvpr}              

\definecolor{cvprblue}{rgb}{0.21,0.49,0.74}
\usepackage[pagebackref,breaklinks,colorlinks,allcolors=cvprblue]{hyperref}

\usepackage{booktabs}
\usepackage{multirow}
\usepackage{graphicx}
\usepackage{algorithm}
\usepackage{algorithmic}
\usepackage{utfsym}

\usepackage{colortbl}

\def\confName{ICCV}
\def\confYear{2025}

\title{
Continual Adaptation: Environment-Conditional Parameter Generation for \\ Object Detection in Dynamic Scenarios
}

\author{Deng Li\textsuperscript{1}, Aming Wu\textsuperscript{2}, Yang Li\textsuperscript{1}, Yaowei Wang\textsuperscript{3}, Yahong Han\textsuperscript{1}\thanks{Corresponding author.}\\
\textsuperscript{1}College of Intelligence and Computing, Tianjin University, Tianjin, China\\
\textsuperscript{2}School of Electronic Engineering, Xidian University, Xi’an, China\\
\textsuperscript{3}Peng Cheng Laboratory, Shenzhen, China\\
{\tt\small lideng@tju.edu.cn, amwu@xidian.edu.cn, liyang1389@tju.edu.cn, wangyw@pcl.ac.cn, yahong@tju.edu.cn}
}

\begin{document}
\maketitle
\begin{abstract}

In practice, environments constantly change over time and space, posing significant challenges for object detectors trained based on a closed-set assumption, i.e., training and test data share the same distribution. To this end, continual test-time adaptation has attracted much attention, aiming to improve detectors' generalization by fine-tuning a few specific parameters, e.g., BatchNorm layers. However, based on a small number of test images, fine-tuning certain parameters may affect the representation ability of other fixed parameters, leading to performance degradation. Instead, we explore a new mechanism, i.e., converting the fine-tuning process to a specific-parameter generation. Particularly, we first design a dual-path LoRA-based domain-aware adapter that disentangles features into domain-invariant and domain-specific components, enabling efficient adaptation. Additionally, a conditional diffusion-based parameter generation mechanism is presented to synthesize the adapter’s parameters based on the current environment, preventing the optimization from getting stuck in local optima. Finally, we propose a class-centered optimal transport alignment method to mitigate catastrophic forgetting. Extensive experiments conducted on various continuous domain adaptive object detection tasks demonstrate the effectiveness. Meanwhile, visualization results show that the representation extracted by the generated parameters can capture more object-related information and strengthen the generalization ability.

\end{abstract}
\vspace{-1.5em}
\section{Introduction}
\label{sec:intro}

Deep learning has achieved remarkable success in various computer vision tasks \cite{he2016deep, rawat2017deep, zou2023object, carion2020end, liu2023bird, liu2024vol, liu2024vision}. 
Most deep learning models are learning under the assumption that the training data and testing data follow the same distribution. However, this assumption often fails to hold in real-world scenarios. In practical scenarios, there is a distribution difference between the training data (source domain) and the test data (target domain), which is commonly referred to as domain shift~\cite{sugiyama2006mixture}.

\begin{figure}[t]
\begin{center}
\includegraphics[width=1 \linewidth]{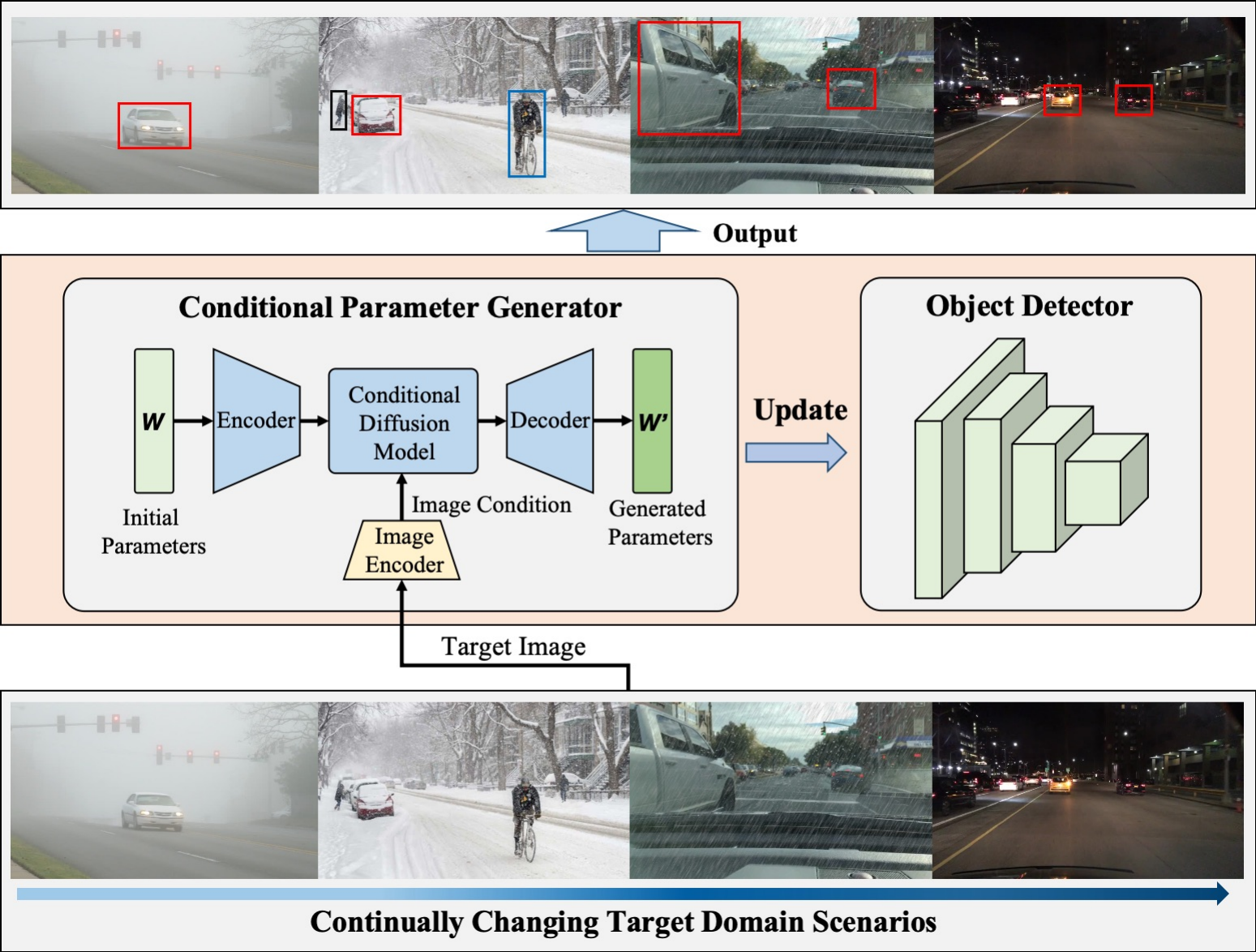}
\end{center}
\vspace{-1.5em}
\caption{\textbf{Illustration of proposed parameter generation method for object detector in continually changing scenarios.} The conditional parameter generator generates robust parameters for the detector with the target scene conditions to improve the detector's generalization performance.
   }
\vspace{-1.5em}
\label{illustration1}
\end{figure}

Domain shift leads to a performance degradation of the model directly tested on the target domains. It is crucial to address this domain shift issue to ensure the robustness and generalization of deep learning models in target domains. To mitigate the effects of domain shift, some domain adaptation methods have been proposed, such as Unsupervised Domain Adaptation (UDA)~\cite{ganin2015unsupervised,murez2018image,sankaranarayanan2018generate,wu2021vector}, Source-Free Domain Adaptation (SFDA)~\cite{kundu2020universal}, Domain Generalization (DG)~\cite{muandet2013domain,ghifary2015domain, carlucci2019domain}, and Test-Time Adaptation (TTA)~\cite{wang2020tent,mirza2023actmad,zhang2022memo}. 
Unlike traditional domain adaptation methods that can access the target domain during training, TTA focuses on adapting the model to the target domain during the test phase without accessing labeled data from the target domain. A more challenging variant of this setting is Continual Test-Time Adaptation (CTTA)~\cite{wang2022continual,song2023ecotta,gan2023cloud}, where the model is required to adapt continually as it encounters dynamic target domain data over time. 
Continual Test-Time Adaptative Object Detection (CTTAOD)~\cite{yoo2024and} expands upon CTTA in the context of object detection, where the detector needs to adapt to the changing data distributions while maintaining high detection accuracy across continually changing environments, making the problem even more complex.

CTTAOD faces several significant challenges. 
\textbf{Challenge 1}: The detector needs to adapt to the changing data distributions with unlabeled test data in an online manner. 
\textbf{Challenge 2}: Computational efficiency is crucial, as the test-time application typically requires rapid adaptation mechanisms.
\textbf{Challenge 3}: The detector needs to ensure stability to avoid catastrophic forgetting. 
Some existing methods have been proposed to address these challenges,
U-VPA \cite{gan2023cloud} proposes a visual prompt adaptation method based on the teacher-student structure to transfer the generalization capability of a large model to the lightweight model.
SKIP~\cite{yoo2024and} introduces a lightweight adapter and two criteria to determine whether to update the adapter.

The solution is to fine-tune a few parameters to strengthen the models' adaptability. However, fine-tuning a few specific parameters may affect other fixed parameters, which may result in the model's optimization getting stuck in local optima due to the continual learning on unlabeled target domains.
To address these limitations and to tackle the aforementioned three challenges, we propose a novel parameter generation method for continual test-time adaptive object detection. 
First, a dual-path LoRA-based domain-aware adapter is proposed to learn both domain-invariant and domain-specific features. During the testing phase, only the parameters of this domain-aware adapter are updated, enabling the detector to adapt efficiently to dynamic target domains. Then, a parameter-generated tuning method is proposed that generates robust parameters for the domain-aware adapter based on the target scene conditions to prevent the model from getting trapped in local optima during continual unsupervised training on the target domain. Finally, to mitigate catastrophic forgetting, we propose a class-centered optimal transport alignment method, which aligns the target domain instance features with the source domain class centers.
The main contributions of our method can be summarized as follows:

(1) We propose a dual-path LoRA-based domain-aware adapter that disentangles features into domain-invariant and domain-specific components, enabling efficient adaptation to continually changing target domain scenarios. 

(2) We propose a conditional diffusion-based parameter generation method to generate robust parameters for the domain-aware adapter and prevent the optimization from becoming stuck in local optima. 

(3) We propose a class-centered optimal transport alignment method to align instance features with source domain class centers to mitigate catastrophic forgetting. 

(4) Extensive experiments conducted on various continual test-time adaptive object detection tasks validate the effectiveness and generality of the proposed method.

\vspace{-1.em}
\section{Related Works}
\label{sec:related_work}

\subsection{Test-Time Adaptation}

Test-Time Adaptation (TTA) aims to adapt a pre-trained model to changing or unseen target domain conditions during the testing phase. Unlike traditional domain adaptation methods, which assume access to both source and target domain data during training \cite{ben2010theory, ganin2016domain}, TTA operates solely during the test phase. 
TENT \cite{wang2020tent} introduced an entropy minimization approach to reduce the uncertainty of the predictions by encouraging confident outputs on test data. It updates the batch normalization by minimizing entropy on test data. Activated by TENT \cite{wang2020tent}, some methods (such as SAR~\cite{niu2023towards}, NORM~\cite{mirza2022norm}, and DUA~\cite{schneider2020improving}) further proposed to explore the generalization capability of normalization layers. 
Some methods~\cite{mirza2023actmad,zhang2022memo} update the full parameters of the model to adapt to the test domain. However, this approach is inefficient for test-time adaptation. 
ActMad~\cite{mirza2023actmad} and MEMO~\cite{zhang2022memo} update the full parameters of the model to adapt to the test domain. However, this approach is inefficient for test-time adaptation. 

Most existing Test-Time Adaptive Object Detection (TTAOD) methods leverage a teacher-student structure to generate pseudo-labels and adapt to the target domain through the Exponential Moving Average (EMA) self-training approach. To mitigate the impact of noisy pseudo-labels, IoU-Filter \cite{ruan2024fully} introduces an Intersection over Union (IoU) filter to obtain higher-quality pseudo-labels. STFAR \cite{chen2023stfar} employs feature distribution alignment as a regularization technique of self-training to eliminate the effect of the noise pseudo-labels. However, due to data augmentation and the computations in both the teacher and student networks, these methods suffer from inefficient parameter updates and slow inference speeds.

\subsection{Continual Test-Time Adaptation}
Many TTA methods encounter limitations when applied to Continual Test-Time Adaptation (CTTA), where models must adapt to a series of domain shifts over time. Additionally, CTTA also faces the challenge of catastrophic forgetting~\cite{kirkpatrick2017overcoming}. To mitigate this, EcoTTA \cite{song2023ecotta} employs a meta-network that regularizes the outputs from both the meta-network and the frozen network. CoTTA \cite{wang2022continual} attempts to address this by incorporating mechanisms like pseudo-labeling and augmentation-based consistency regularization, but it often struggles to maintain stable performance across continually changing domains. U-VPA \cite{gan2023cloud} proposes a visual prompt adaptation method to transfer the generalization knowledge of a large model to a lightweight detector. SKIP~\cite{yoo2024and} introduces a lightweight adapter module and proposes two criteria to determine dynamic skipping for continual test-time adaptive object detection.

\subsection{Parameter Generation}
A parameter generation model is a generative model designed to generate parameters for other neural networks \cite{ha2017hypernetworks}. HyperStyle \cite{alaluf2022hyperstyle} learns the weights of StyleGAN modules using a hypernetwork. Peebles et al. \cite{peebles2022learning} collect model checkpoints to construct a parameter dataset and train a parameter generation model with transformer diffusion. HyperDiffusion \cite{erkoc2023hyperdiffusion} trains a diffusion model to generate neural implicit fields by learning the weights of an MLP. GPD \cite{yuan2024spatio} introduces a generative pre-training framework to adaptively generate unique parameters for spatio-temporal prediction models.
Unlike the aforementioned methods, we propose a conditional diffusion parameter generation model that generates LoRA-based adapter parameters with the condition of domain image features for continual test-time adaptation in object detection tasks.

\vspace{-1.em}
\section{Methodology}

\begin{figure*}[]
\begin{center}
\includegraphics[width=0.9 \linewidth]{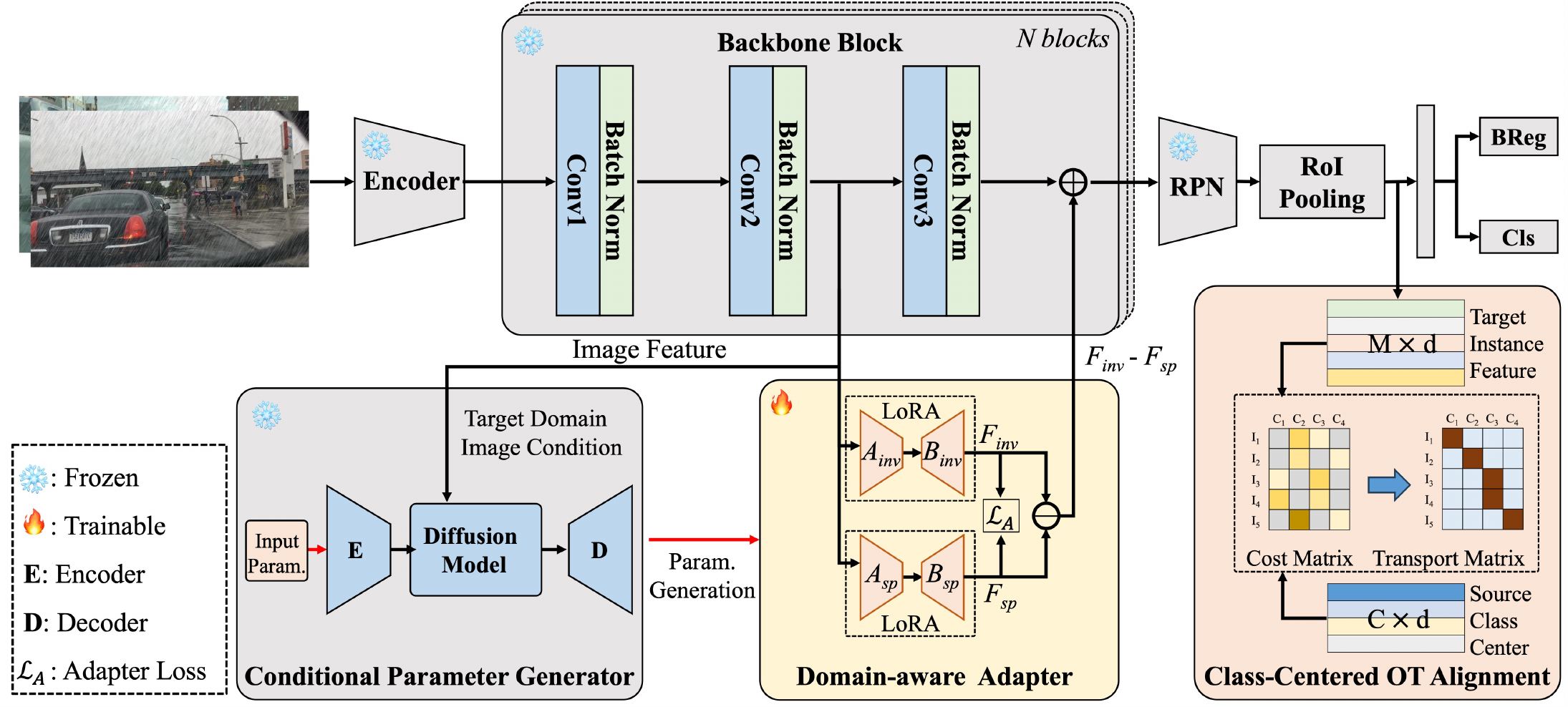}
\end{center}
\vspace{-2em}
\caption{\textbf{Illustration of our proposed parameter-generated adapter for continual test-time adaptive object detection.} 
This method mainly includes the conditional parameter generator, domain-aware adapter, and class-centered optimal transport alignment module. 
The domain-aware adapter is introduced into multiple blocks of the backbone. It disentangles the features into domain-invariant features \( F_{inv} \) and domain-specific features \( F_{sp} \). 
The class-centered optimal transport alignment module is used to mitigate the issue of catastrophic forgetting.
The conditional parameter generator is used to generate the parameters of the adapter with scene information to prevent the optimization from getting stuck in local optima.
   }
\vspace{-1.5em}
\label{arch}
\end{figure*}

\vspace{-0.5em}
\subsection{Problem Formulation}
Continual Test-Time Adaptive Object Detection (CTTAOD) aims to adapt a pre-trained object detection model $f_\theta$ to the dynamic scenarios at test time. 
The model is initially trained on a labeled \emph{source domain} $\mathcal{D}_S = \{(x_i^S, y_i^S)\}_{i=1}^{N_S}$. Here, $x_i^S$ represents the input image, and $y_i^S$ represents the corresponding ground-truth annotations for object detection (e.g., bounding boxes and class labels). At test phase, the model is continual tested and adapted on the \emph{unlabeled target domain} $\mathcal{D}_T = \{x_i^T\}_{i=1}^{N_T}$ which consists of a sequence of unlabeled images $x_i^T$, spanning multiple domains with varying degrees of domain shift. The goal is to update the model parameters $\theta$ online with the test data stream to maintain robust detection performance.

\vspace{-0.5em}
\subsection{Framework}

To improve the generalization ability of the model in continually changing environments, we propose a parameter-generated tuning method for continual test-time adaptive object detection, as shown in Fig.~\ref{arch}. 
The proposed approach consists of three key components. First, the LoRA-based domain-aware adapter disentangles features into domain-invariant and domain-specific components.
By leveraging domain-invariant features for prediction while minimizing dependence on domain-specific features, the model achieves improved generalization across diverse domains. Additionally, it adapts to dynamic environments by updating only the adapter parameters, ensuring efficient and flexible adaptation without modifying the entire model.
Second, the pre-trained conditional diffusion-based parameter generator generates robust parameters for the domain-aware adapter conditioned on the target domain scene. 
Due to the diversity of the generated parameters, it helps prevent getting stuck in local optima within the parameter space, leading to more generalizable parameters.
Finally, the class-centered optimal transport alignment module aligns the target instance features with the source class-centered features to eliminate catastrophic forgetting.

\vspace{-0.5em}
\subsection{LoRA-based Domain-aware Adapter}

The primary challenge in test-time domain adaptation is adapting a model trained on a source domain to perform effectively on a target domain with a domain shift. To address this, we propose a domain-aware adapter that disentangles features into domain-invariant and domain-specific components. This enables the model to consistently recognize objects across different domains while flexibly adapting to domain-specific characteristics, thereby enhancing its generalization and robustness in dynamic scenarios.

Inspired by the LoRA \cite{hu2022lora}, which enables efficient fine-tuning through low-rank matrices, we propose a dual-path LoRA-based adapter in multiple blocks of the backbone for learning domain-invariant features \( F_{inv} \) and domain-specific features \( F_{sp} \). Finally, the difference \( F_{inv}-F_{sp} \)) between the domain-invariant features \( F_{inv} \) and the domain-specific features \( F_{sp} \) is fed into the backbone.
The updated weight matrix \( W_{up} \) can be written as:
\begin{equation}
W_{up} = W_{b} + \Delta W_{inv} - \Delta W_{sp} \\ = W_{b} + A_{inv}B_{inv}-A_{sp} B_{sp},
\end{equation}
where \( A_{inv} \in \mathbb{R}^{d \times r_1} \) and \( A_{sp} \in \mathbb{R}^{d \times r_2} \)  are the learnable matrix that projects the input into a lower-dimensional space.  \( B_{inv} \in \mathbb{R}^{r_1 \times d} \) and \( B_{sp} \in \mathbb{R}^{r_1 \times d} \) are the learnable matrix project the low-rank features back into the original space. \( r_1 \) and \( r_2 \) are the rank of the low-rank approximation for the domain-invariant and domain-specific features, respectively. \( d \) is the dimensionality of the weight matrix.

To ensure that domain-invariant features and domain-specific features remain \textit{distinct} and \textit{statistically independent}, we introduce the Orthogonal Loss and Hilbert-Schmidt Independence Criterion (HSIC) loss. 
The orthogonal loss is defined as:s
\begin{equation}
\mathcal{L}_{orth} = \left\| F_{inv}^T F_{sp} \right\|_F^2,
\end{equation}
where $\| \cdot \|_F$ denotes the Frobenius norm. Minimizing this loss encourages the two feature spaces to be orthogonal, ensuring that domain-invariant and domain-specific features capture distinct information.
HSIC measures the dependence between two random variables based on the Hilbert-Schmidt norm of their cross-covariance operator.
Given two sets of features $F_{inv}$ and $F_{sp}$, the HSIC loss is computed as:
\begin{equation}
\mathcal{L}_{HSIC} = HSIC(F_{inv}, F_{sp}),
\end{equation}
and the function of HSIC can be approximated as:
\begin{equation}
HSIC(F_{inv}, F_{sp}) = \frac{1}{(n-1)^2} Tr(K_{inv} H K_{sp} H),
\end{equation}
where $K_{inv}$ and $K_{sp}$ are the Gram matrices corresponding to $F_{inv}$ and $F_{sp}$, and $H = I_n - \frac{1}{n} \mathbf{1}_n \mathbf{1}_n^T$ is the centering matrix. Minimizing the HSIC loss reduces the statistical dependence between the domain-invariant and domain-specific features.

The loss function of the proposed domain-aware adapter can be written as:
\begin{equation}
\mathcal{L}_{A} = \lambda_{orth} \mathcal{L}_{orth} + \lambda_{HSIC} \mathcal{L}_{HSIC}.
\end{equation}

\vspace{-0.5em}
\subsection{Conditional Parameter Generation}

We propose a parameter generator based on a conditional diffusion model to generate the robust parameters for the domain-aware adapter. The motivation is to prevent the adapter parameters from getting stuck in local optima during the unsupervised learning process, thereby enhancing the generalization of the adapter. To enable efficient updating of the adapter parameters during the test phase, we first train the designed adapter and the conditional parameter generator together with the detector on source domain data. This training process allows the generator to learn robust parameter generation capabilities, a stage we refer to as offline training. During the test phase, the conditional parameter generator is frozen and only used for inference to generate the robust parameters of the domain-aware adapter.

As illustrated in Fig.~\ref{arch}, our proposed parameter generation module consists of a conditional diffusion model and an autoencoder module.
The autoencoder is used to compress the high-dimensional adapter parameter matrices into low-dimensional latent vectors and reconstruct the original parameter matrices from these latent vectors. It learns a latent representation of the parameters, capturing their distribution and essential characteristics. 
The autoencoder consists of two parts: an encoder \( \mathbf{E}: \mathbb{R}^{d \times d} \rightarrow \mathbb{R}^z \), which maps the parameters to a lower-dimensional latent space, and a decoder \( \mathbf{D}: \mathbb{R}^z \rightarrow \mathbb{R}^{d \times d} \), which reconstructs the parameter matrix from the latent space vector. It is trained with the reconstruction loss \( \mathcal{L}_{recon} \), which ensures that the decoded parameters \( \mathbf{D}(\mathbf{E}(W_{Adapter})) \) are as close as possible to the input adapter parameters \( W_{Adapter} \):
\begin{equation}
\mathcal{L}_{recon} = \left\| W_{Adapter} - \mathbf{D}(\mathbf{E}(W_{Adapter})) \right\|_F^2.
\end{equation}

\textbf{Diffusion Process.}
The diffusion model follows a forward process that gradually adds Gaussian noise to the latent representation of the parameters over a series of time steps and a reverse process that removes noise step by step, conditioned on the target domain image features. Let \( z_0 = \mathbf{E}(W_{Adapter}) \) be the latent vector obtained from the encoder. We define a sequence of latent vectors \( \{ z_t \}_{t=1}^T \), where \( z_T \) is a pure noise vector and \( z_0 \) corresponds to the clean latent vector. The forward process adds noise to \( z_0 \) at each step \( t \), following a Gaussian distribution:
\begin{equation}
q(z_t | z_{t-1}) = \mathcal{N}(z_t; \sqrt{1 - \beta_t} z_{t-1}, \beta_t I),
\end{equation}
where \( \beta_t \) controls the variance of the noise added at step \( t \), and \( I \) is the identity matrix.

\textbf{Reverse Process and Conditional Denoising.}
The reverse process aims to denoise the noisy latent vector \( z_T \) back to a clean latent vector \( z_0 \). This process is conditioned on the target domain image features \( x_T \), which provide the necessary domain-specific information for generating adapter parameters suited to the target domain. The denoising process is modeled by a neural network \( \epsilon_\theta(z_t, x_T, t) \), which predicts the noise added at each step \( t \).
Then, the predicted noise is subtracted from the noisy latent vector at each step to progressively recover the clean latent vector.

The loss function for training the diffusion model is the denoising score matching loss, which minimizes the difference between the predicted noise and the actual noise added during the forward process:
\begin{equation}
\mathcal{L}_{diff} = \mathbb{E}_{z_0, t, \epsilon} \left[ \left\| \epsilon - \epsilon_\theta(z_t, x_T, t) \right\|_2^2 \right],
\end{equation}
where \( \epsilon \) is the sampled noise from the Gaussian distribution.

\textbf{Inference: Robust Adapter Parameters Generation.}
In our method, the conditional parameter generator is pre-trained with the detector on the source domain data in an offline way. 
In the test phase, we freeze the parameters of the conditional parameter generator and generate robust parameters for the domain-aware adapter conditioned on the target domain images. 
At each step during the test-time adaptation, the adapter parameters \( W_{Adapter} \) are updated with the conditional parameter generator. 
The parameter generator takes the current target domain frame \( x_T^t \) and the trained adapter parameters \( W_{Adapter} \) as input to generate robust parameters \( W_{Adapter}^t \):

\begin{equation}
W_{Adapter}^t = \mathbf{D}(\mathbf{Diff}( x_T^t, W_{{Adapter}})).
\end{equation}

The reconstructed parameters \( W_{Adapter} \) are used for the domain-ware adapters, ensuring that the model can adapt effectively to the new domain without falling into local optima, thus enhancing the generalization performance of detectors in continuous test-time adaptation.

\vspace{-0.5em}
\subsection{Class-Centered Optimal Transport Alignment}
To mitigate the catastrophic forgetting issue, we propose a class-centered optimal transport alignment method that preserves source domain knowledge by aligning the target domain instances with the class centers of the source domain through the Optimal Transport (OT) method. It ensures that the model retains the structural characteristics of the source domain while adapting to the new distribution of the target domain.
Let \( \{ x_{t,j}^c \}_{j=1}^{N_t^c} \) denote the $j$-th instance in class \( c \) in the target domain, where \( N_t^c \) is the number of target domain instances in class \( c \). The transportation cost from a target instance \( x_{t,j}^c \) to the source domain class center \( \mu_s^c \) is typically defined as the squared Euclidean distance:

\begin{equation}
\mathbf{C}(x_{t,j}^c, \mu_s^c) = \| x_{t,j}^c - \mu_s^c \|_2^2.
\end{equation}
The goal is to find a transport plan \( \mathbf{P}^c \) that minimizes the total cost of transporting all target instances in class \( c \) to the source class center \( \mu_s^c \). This is formulated as the following optimal transport problem:

\begin{equation}
\min_{\mathbf{P}^c} \sum_{j=1}^{N_t^c} \mathbf{P}_j^c C(x_{t,j}^c, \mu_s^c),
\end{equation}
where \( \mathbf{P}_j^c \) represents the transport plan, and it should satisfy the marginal constraints, ensuring that all target domain instances are mapped to the source domain class centers.
The overall optimal transport loss for aligning the target domain instances to the source domain class centers is defined as:
\begin{equation}
\mathcal{L}_{OT} = \sum_{c=1}^{C} \sum_{j=1}^{N_t^c} \mathbf{P}_j^c \mathbf{C}(x_{t,j}^c, \mu_s^c),
\end{equation}
where \( \mathbf{C} \) is the transportation cost function, and \( C \) denotes the number of classes. This loss function captures the total cost of transporting all target domain instances to their corresponding source domain class centers.

\vspace{-0.5em}
\subsection{Total Loss Function in Test Phase}
Since the conditional parameter generator is pre-trained offline on source domain data, the total loss function during the test phase includes the loss functions of both the domain-aware adapter and the class-centered alignment module, which can be expressed as:

\begin{equation}
\mathcal{L}_{\mathrm{total}}=\lambda_{A}\mathcal{L}_{A}+\lambda_{CA}\mathcal{L}_{OT},
\label{total_loss}  
\end{equation}
where $ \lambda_{A} $  and $\lambda_{CA}$ are the weights of the loss function of the domain-aware adapter and class-centered alignment module, respectively.

At each optimization step, the conditional parameter generator takes the optimized adapter parameters and the current image as inputs to generate new robust parameters for the domain-aware adapter.

\vspace{-0.5em}
\section{Experiments}

\vspace{-0.5em}
\subsection{Datasets}

\textbf{Diverse-Weather}~\cite{wu2022single} is a urban-scene detection domain generalization benchmark. The dataset consists of five domains with varying weather conditions: Daytime Clear, Night Clear, Dusk Rainy, Night Rainy, and Daytime Foggy. We set the Daytime Clear as the source domain, containing 19,395 training images. The remaining four domains are set as target domains, with the Night Clear domain containing 26,158 images, the Dusk Rainy domain containing 3,501 images, the Night Rainy domain containing 2,494 images, and the Daytime Foggy domain containing 3,775 images.

\textbf{COCO} \cite{lin2014microsoft} is one of the most widely used benchmarks for object detection. It provides a large-scale, richly annotated dataset containing real-world images with complex scenes. This dataset contains 80 categories. The training set contains 118,000 images, and the validation set contains 5,000 images. \textbf{COCO-C} dataset refers to a corrupted version of the COCO dataset, which is commonly used for evaluating the robustness of object detection models. 
It introduces various corruptions~\cite{michaelis2019benchmarking}, such as noise (e.g., Gaussian, shot noise), blur (e.g., defocus, motion), weather effects (e.g., snow, fog), and digital distortions (e.g., JPEG compression, pixelation) to simulate real-world degradation in images. These corruptions are applied at multiple severity levels, typically ranging from mild to extreme, allowing for the evaluation of model robustness under different challenging conditions.

\textbf{SHIFT}~\cite{sun2022shift} is a multi-task synthetic dataset for autonomous driving aimed at studying the adaptability and robustness of perception systems in continually changing environments. The dataset contains 6 categories and also includes 5 weather conditions: Clear, Cloudy, Overcast, Foggy, and Rainy, as well as 3 different time conditions: Daytime, Dawn/Dusk, and Night, simulating the transition from day to night. 

\vspace{-0.5em}
\subsection{Implementation Details}
Consistent with other continual Test-Time Adaptation for Object Detection (TTAOD) methods, we conducted our experiments using the Faster R-CNN framework. To validate the effectiveness of the proposed method, we evaluated it on three domain-adaptive datasets with dynamically changing target domains. Specifically, for the Diverse-Weather Dataset, which simulates weather variations in urban road scenarios, we set the backbone to ResNet-101. For the COCO and SHIFT datasets, the backbones were set as ResNet-50~\citep{he2016deep} and Swin-T~\cite{liu2021swin}, respectively. We used SGD with a weight decay of 0.0005 as the optimizer, set the learning rate to 0.0001, and employed a batch size of 4 during training. Following the SKIP~\cite{yoo2024and}, we set the severity of corruption as 5 for COCO-C.

\vspace{-0.5em}
\subsection{Experimental Results and Analysis}
\textbf{Compared Methods.} 
For the Diverse-Weather dataset, we compared our method with several state-of-the-art approaches, including SKIP~\cite{yoo2024and}, which introduces architecture-agnostic and lightweight adaptor modules and proposes two criteria to determine dynamic skipping adaptation. Additionally, we compared our method with several domain generalization methods, such as SW~\cite{pan2019switchable}, IBN-Net~\cite{pan2018two}, IterNorm~\cite{huang2019iterative}, and ISW~\cite{choi2021robustnet}. Specifically, SW~\cite{pan2019switchable} allows the model to dynamically switch between different whitening operations, ISW~\cite{choi2021robustnet} selects relevant instances for whitening, IBN-Net~\cite{pan2018two} combines instance normalization and batch normalization, and IterNorm~\cite{huang2019iterative} introduces an iterative approach for normalization.

\begin{table}[]
\setlength{\tabcolsep}{2pt}
\begin{center}
\begin{tabular}{l|c|cccc|c}
\toprule[1pt]
Method  & Pub.   &\multicolumn{1}{c}{\begin{tabular}[c]{@{}c@{}}Dusk\\ Rainy\end{tabular}}   &\multicolumn{1}{c}{\begin{tabular}[c]{@{}c@{}}Daytime\\ Foggy\end{tabular}}  &\multicolumn{1}{c}{\begin{tabular}[c]{@{}c@{}}Night\\ Rainy \end{tabular}} &\multicolumn{1}{c}{\begin{tabular}[c]{@{}c@{}} Night \\ Clear\end{tabular}} &\multicolumn{1}{|c}Avg.\\ 
\hline
Direct-Test   &-  & 29.6 & 33.4 & 13.2 & \textbf{35.5}& 27.9  \\
IBN-Net  &ECCV'18  & 26.1 & 29.6   & 14.3  & 32.1&25.5  \\
IterNorm  & CVPR'19& 22.8 &28.4 & 12.6 & 29.6 &23.4   \\
SW  &ICCV'19  & 26.3  & 30.8 & 13.7 & 33.4 & 26.1 \\ 
ISW &CVPR'21 & 25.9 & 31.8  & 14.1 & 33.2&  26.3  \\
\hline   
SKIP  &CVPR'24  & 34.1& 32.9& 17.3&32.6&29.2\\
\cellcolor[rgb]{ .851,  .851,  .851}Ours  &\cellcolor[rgb]{ .851,  .851,  .851}- &\cellcolor[rgb]{ .851,  .851,  .851}\textbf{35.7}&\cellcolor[rgb]{ .851,  .851,  .851}\textbf{33.7}&\cellcolor[rgb]{ .851,  .851,  .851}\textbf{18.2}&\cellcolor[rgb]{ .851,  .851,  .851} 35.3 &\cellcolor[rgb]{ .851,  .851,  .851}\textbf{30.7}    \\ 
\bottomrule[1pt]
\end{tabular}
\end{center}
\vspace{-1.5em}
\caption{Continual test-time adaptive object detection results (\%) on Diverse-Weather with the backbone of ResNet-101. The Day-Clear domain is set as the source domain.
}
\vspace{-2em}
\label{result_diverse_weather}
\end{table}

\begin{table*}[ht]
\centering
\resizebox{\textwidth}{!}{%
\begin{tabular}{ll|ccccccccccccccc|c|c}
\toprule[1pt]
\multirow{2}{*}{Backbone} & \multirow{2}{*}{Method} & \multicolumn{3}{c}{Noise} & \multicolumn{4}{c}{Blur} & \multicolumn{4}{c}{Weather} & \multicolumn{4}{c|}{Digital} & \multirow{2}{*}{Source} &  \multirow{2}{*}{Avg.}  \\
\cmidrule(lr){3-5} \cmidrule(lr){6-9} \cmidrule(lr){10-13} \cmidrule(lr){14-17}
 & & Gau & Sht & Imp & Def & Gls & Mtn & Zm & Snw & Frs & Fog & Brt & Cnt & Els & Px & Jpg & & \\
\midrule

\multirow{7}{*}{ResNet-50} & Direct-Test & 9.1 & 11.0 & 9.8 & 12.6 & 4.5 & 8.8 & 4.6 & 19.1 & 23.1 & 38.4 & 38.0 & 21.4 & 15.6 & 5.3 & 11.9 & \textbf{44.2}& 17.3 \\
 & NORM & 9.9 & 11.9 & 11.0 & 12.6 & 5.2 & 9.1 & 5.1 & 19.4 & 23.5 & 38.2 & 37.6 & 22.4 & 17.2 & 5.7 & 10.3 & 43.4 & 17.5  \\
 & DUA & 9.8 & 11.7 & 10.8 & 12.8 & 5.2 & 8.9 & 5.1 & 19.3 & 23.7 & 38.4 & 37.8 & 22.3 & 17.2 & 5.4 & 10.1 & 44.1 & 17.1\\
 & ActMAD & 9.1 & 9.6 & 7.0 & 11.0 & 3.2 & 6.1 & 3.3 & 12.8 & 14.0 & 27.7 & 27.8 & 3.9 & 12.9 & 2.3 & 7.2 & 34.3 & 10.5 \\
 & Mean-Teacher & 9.6 & 12.5 & 12.0 & 4.0 & 2.9 & 4.8 & 3.1 & 16.2 & 23.5 & 35.1 & 34.0 & 21.8 & 16.6 & 8.2 & 12.7 & 40.3 & 14.5  \\
  & SKIP  & 12.7 & 17.8 & 17.5 & 12.4 & 11.5 & 11.3 & 6.6 & 22.8 & \textbf{26.9} & \textbf{38.6} & \textbf{38.5} & \textbf{28.0} & 25.1 & 21.2 & 22.2 & 41.8 & 22.2  \\

  &\cellcolor[rgb]{ .851,  .851,  .851} Ours  &\cellcolor[rgb]{ .851,  .851,  .851}  \textbf{15.4} &\cellcolor[rgb]{ .851,  .851,  .851} \textbf{19.7} &\cellcolor[rgb]{ .851,  .851,  .851} \textbf{19.3} &\cellcolor[rgb]{ .851,  .851,  .851} \textbf{13.1} &\cellcolor[rgb]{ .851,  .851,  .851} \textbf{12.3} &\cellcolor[rgb]{ .851,  .851,  .851} \textbf{12.1} &\cellcolor[rgb]{ .851,  .851,  .851} \textbf{6.9} &\cellcolor[rgb]{ .851,  .851,  .851} \textbf{22.9} &\cellcolor[rgb]{ .851,  .851,  .851} 26.7 &\cellcolor[rgb]{ .851,  .851,  .851} 38.2 &\cellcolor[rgb]{ .851,  .851,  .851} 38.3 &\cellcolor[rgb]{ .851,  .851,  .851} 27.2 &\cellcolor[rgb]{ .851,  .851,  .851} \textbf{25.3} &\cellcolor[rgb]{ .851,  .851,  .851} \textbf{24.1} & \cellcolor[rgb]{ .851,  .851,  .851}\textbf{23.2} & \cellcolor[rgb]{ .851,  .851,  .851}42.4 & \cellcolor[rgb]{ .851,  .851,  .851}\textbf{23.0}\\
  \hline
 \multirow{5}{*}{SwinT} & Direct-Test & 9.7 & 11.4 & 10.0 & 13.4 & 7.5 & 12.1 & 5.2 & 20.7 & 24.8 & 36.1 & 36.0 & 12.9 & 19.1 & 4.9 & 15.8 & \textbf{43.0} & 17.7 \\
 & ActMAD & 10.7 & 12.0 & 9.4 & 12.3 & 5.7 & 9.5 & 4.5 & 15.3 & 17.5 & 27.6 & 28.2 & 1.1 & 16.7 & 2.6 & 8.7 & 36.3 & 13.9 \\
 & Mean-Teacher & 10.0 & 12.1 & 11.2 & 12.8 & 8.1 & 12.1 & 4.9 & 19.6 & 23.7 & 34.9 & 34.0 & 8.0 & 18.9 & 6.1 & 17.6 & 41.0 & 17.2 \\
  & SKIP  & 13.3 & 15.3 & 15.1 & 14.0 & 12.8 & 13.9 & 6.5 & 22.0 & 25.4 & 35.5 & 34.9 & \textbf{26.5} & 25.9 & \textbf{23.4} & 20.2 & 41.2 & 21.6 \\
  &\cellcolor[rgb]{ .851,  .851,  .851} Ours  &\cellcolor[rgb]{ .851,  .851,  .851}\textbf{14.2} &\cellcolor[rgb]{ .851,  .851,  .851} \textbf{16.9} &\cellcolor[rgb]{ .851,  .851,  .851} \textbf{16.5} & \cellcolor[rgb]{ .851,  .851,  .851}\textbf{14.4} &\cellcolor[rgb]{ .851,  .851,  .851} \textbf{13.3} &\cellcolor[rgb]{ .851,  .851,  .851} \textbf{14.2} &\cellcolor[rgb]{ .851,  .851,  .851} \textbf{8.1} &\cellcolor[rgb]{ .851,  .851,  .851} \textbf{23.3} &\cellcolor[rgb]{ .851,  .851,  .851} \textbf{26.9} &\cellcolor[rgb]{ .851,  .851,  .851} \textbf{37.1} & \cellcolor[rgb]{ .851,  .851,  .851}\textbf{36.8} & \cellcolor[rgb]{ .851,  .851,  .851}26.4 & \cellcolor[rgb]{ .851,  .851,  .851}\textbf{26.4} &\cellcolor[rgb]{ .851,  .851,  .851} 20.7 &\cellcolor[rgb]{ .851,  .851,  .851} \textbf{21.7} &\cellcolor[rgb]{ .851,  .851,  .851} 42.3& \cellcolor[rgb]{ .851,  .851,  .851}\textbf{22.5}\\

\bottomrule[1pt]
\end{tabular}%
}
\vspace{-1.em}
\caption{Continual test-time adaptive object detection results (\%) on COCO $\rightarrow$ COCO-C with the backbone of ResNet-50 and Swin-T.
}
\label{result_coco}
\vspace{-1.em}
\end{table*}

\begin{figure*}[t]
\begin{center}
\includegraphics[width=0.91 \linewidth]{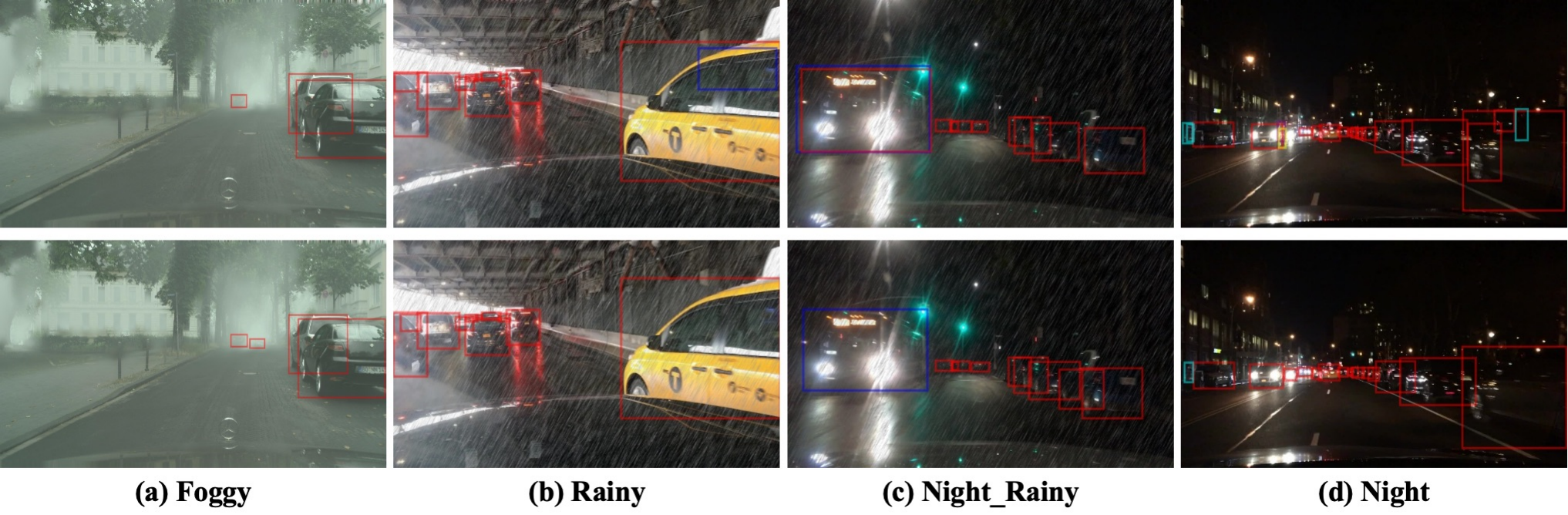}
\end{center}
\vspace{-1.8em}
\caption{\textbf{Detection results of the continual test-time adaptation on the urban scene Diverse-Weather Dataset~\cite{wu2022single}}, where the top row represents the detection results of SKIP~\cite{yoo2024and}, and the bottom row corresponds to our proposed method. 
In the ``Daytime Foggy" scene, our method accurately detects small-sized {\color[rgb]{1.00,0.00,0.00} cars}. In the ``Dusk Rainy" scene, SKIP~\cite{yoo2024and} incorrectly detects an object as a {\color[rgb]{0.00,0.00,1.00} bus}. In the complex ``Night Rainy" scene, SKIP~\cite{yoo2024and} misclassifies a bus as a {\color[rgb]{1.00,0.00,0.00} car}, while our method accurately identifies the object as a {\color[rgb]{0.00,0.00,1.00} bus}.  Furthermore, in the ``Night Clear" scene, SKIP~\cite{yoo2024and} misclassifies an object as a {\color[rgb]{0.00,1.00,1.00} person}, while our method accurately locates the {\color[rgb]{1.00,0.00,0.00} cars}.  
   }
\vspace{-1.em}
\label{visual_box}
\end{figure*}

For the COCO and SHIFT datasets, we compared our method with several state-of-the-art Test-Time Adaptation (TTA) approaches, including NORM~\cite{schneider2020improving}, DUA ~\cite{mirza2022norm}, ActMAD~\cite{mirza2023actmad}, and SKIP~\cite{yoo2024and}. Specifically, NORM~\cite{schneider2020improving} and DUA~\cite{mirza2022norm} address domain shifts by updating only the parameters of the batch normalization (BN) layers, while ActMAD~\cite{mirza2023actmad} aligns the distribution of output features across all BN layers. Additionally, we compared our approach with the Mean-Teacher method, which is based on the student-teacher structure and updates all parameters of the student model.

\begin{table*}[ht]
\centering
\resizebox{\textwidth}{!}{%
\begin{tabular}{ll|ccccccc|c|cc|c}
\toprule[1pt]
\multirow{2}{*}{Backbone} & \multirow{2}{*}{Method} & \multicolumn{8}{c}{Discrete} & \multicolumn{3}{|c}{Continuous}  \\
\cmidrule(lr){3-10} \cmidrule(lr){11-13} 
 & &cloudy	&overc.	&fog&	rain&		dawn&	night& clear & Avg. & clear$\leftrightarrow$fog & clear$\leftrightarrow$rain & Avg.  \\
\midrule

\multirow{7}{*}{ResNet-50} & Direct-Test &49.4 & 37.9 & 19.7 & 43.1 & 20.1 & 35.3 & \textbf{45.6}  & 35.9 & 12.1  & 15.4& 13.8  \\
 & NORM & 49.7 & 38.6 & 22.9 & 44.7 & 25.1 & 37.4 & 45.5 & 37.7 &16.9 &19.4& 18.2 \\
 & DUA & 45.2 & 31.5 & \textbf{27.7} & 31.9 & 15.2 & 18.6 & 21.1 & 27.3 &\textbf{22.5} &22.4& \textbf{22.5}\\
 & ActMAD & 49.2 & 37.7 & 18.0 & 40.6 & 16.0 & 32.9 & 44.3 & 34.1 & 12.7 & 16.3& 14.5\\
& Mean-Teacher &49.6 & 38.4 & 26.8 & 43.4 & 26.6 & 33.1 & 41.6 & 37.1 & 16.0 & 20.8& 18.4  \\
& SKIP  & 49.7 & 38.8 & 26.9 & 46.2 & \textbf{27.6} & 38.8 & 45.0 & 39.0 & 20.0 & 22.5 &21.2 \\
& \cellcolor[rgb]{ .851,  .851,  .851}Ours  &\cellcolor[rgb]{ .851,  .851,  .851}\textbf{50.2} &\cellcolor[rgb]{ .851,  .851,  .851} \textbf{39.3} &\cellcolor[rgb]{ .851,  .851,  .851} 27.6 &\cellcolor[rgb]{ .851,  .851,  .851} \textbf{46.8} &\cellcolor[rgb]{ .851,  .851,  .851} \textbf{27.6} &\cellcolor[rgb]{ .851,  .851,  .851} \textbf{39.4} &\cellcolor[rgb]{ .851,  .851,  .851} 45.3&\cellcolor[rgb]{ .851,  .851,  .851} \textbf{39.5} &\cellcolor[rgb]{ .851,  .851,  .851} 21.2 &\cellcolor[rgb]{ .851,  .851,  .851} \textbf{23.1} &\cellcolor[rgb]{ .851,  .851,  .851} 22.2 \\  
\hline
\multirow{5}{*}{SwinT} & Direct-Test & 50.0 & 38.9 & 23.1& 45.1 & 26.9 & 39.5 & 45.9 & 38.5  & 18.1 & 21.1&19.6 \\
& ActMAD & 49.8 & 38.4 & 21.4 & 43.1 & 19.0 & 32.0 & 44.8 & 35.5   & 15.6 & 16.3& 16.0\\
& Mean-Teacher & 50.0 & 39.2 & 25.7 & 45.4 & 26.0 & 37.5 & 42.2 & 38.0 & 20.4 & \textbf{24.3} &22.4\\
& SKIP  & 50.3 & 39.7 & 29.1 & 47.1 & 30.2 & 41.5 & 45.9 & 40.6 & 25.1 & 23.8& 24.5\\
&\cellcolor[rgb]{ .851,  .851,  .851} Ours  &\cellcolor[rgb]{ .851,  .851,  .851}\textbf{50.6} &\cellcolor[rgb]{ .851,  .851,  .851} \textbf{39.9} &\cellcolor[rgb]{ .851,  .851,  .851} \textbf{29.7} &\cellcolor[rgb]{ .851,  .851,  .851} \textbf{47.9} &\cellcolor[rgb]{ .851,  .851,  .851} \textbf{30.7} &\cellcolor[rgb]{ .851,  .851,  .851} \textbf{41.9} &\cellcolor[rgb]{ .851,  .851,  .851} \textbf{46.1} &\cellcolor[rgb]{ .851,  .851,  .851} \textbf{41.0} &\cellcolor[rgb]{ .851,  .851,  .851} \textbf{25.7} &\cellcolor[rgb]{ .851,  .851,  .851} 24.1&\cellcolor[rgb]{ .851,  .851,  .851}\textbf{24.9} \\
\bottomrule[1pt]

\end{tabular}%
}
\vspace{-1.em}
\caption{Continual test-time adaptive object detection results (\%) on SHIFT with the backbone of ResNet-50 and Swin-T.
}
\label{result_shift}
\vspace{-1.em}
\end{table*}

\textbf{Diverse-Weather.} 
Table~\ref{result_diverse_weather} shows the experimental results on the Diverse-Weather dataset, where Dusk-Rainy, Daytime-Foggy, Night-Rainy, and Night-Clear are set as the continual target domain scenarios. Our method achieves the best average performance of 30.7\%, outperforming the SOTA method SKIP by 1.5\% (30.7\% - 29.2\%). Notably, our approach achieves the best results on three target domains: Dusk-Rainy, Daytime-Foggy, and Night-Rainy. 
Our method maintains the comparable performance on the Night-Clear domain to the Direct-Test method. However, SKIP suffers significant performance degradation on this domain.
These results validate the effectiveness of our proposed approach in diverse weather scenarios.

\textbf{COCO $\rightarrow$ COCO-C.} 
Table~\ref{result_coco} shows the experimental results on the scenes of COCO $\rightarrow$ COCO-C. The target domain scenarios are organized sequentially based on the types of corruption: Noise, Blur, Weather, and Digital. We conducted experiments using both CNN-based ResNet-50 and Transformer-based Swin-T as the backbone networks.
As shown, our method achieves the best average performance, and outperforms the SKIP by 0.8\% (23.0\%-22.2\%) and 0.9\% (22.5\% - 21.6\%) with the backbone of ResNet-50 and Swin-T, respectively. 
After adapting to COCO-C, we also evaluated the impact of catastrophic forgetting on the validation set of the source domain. The results show that our method achieves 42.4\% and 42.3\% on the source domain using ResNet-50 and Swin-T as the backbone, respectively. 
These experimental results demonstrate the generalization and mitigating catastrophic forgetting of our proposed method in various corrupted scenarios.

\textbf{SHIFT.} 
Table~\ref{result_shift} shows the experimental results on the SHIFT dataset, where we conducted continual test-time adaptation experiments in both the SHIFT-Discrete and SHIFT-Continuous scenarios.
In the SHIFT-Discrete scenario, our proposed method achieved the best average performance and outperforms the SKIP method by 0.5\% (39.5\% - 39.0\%) and 1.0\% (41.0\%  - 30.0\%) with ResNet-50 and Swin-T as the backbone, respectively. In the SHIFT-Continuous scenario, our proposed method also achieved the highest average performance, with 22.2\% and 24.9\% with ResNet-50 and Swin-T as the backbone, respectively. These results demonstrate the effectiveness of our proposed method in the synthetic 
 autonomous driving scenarios.

\begin{figure*}[]
\begin{center}
\includegraphics[width=0.92  \linewidth]{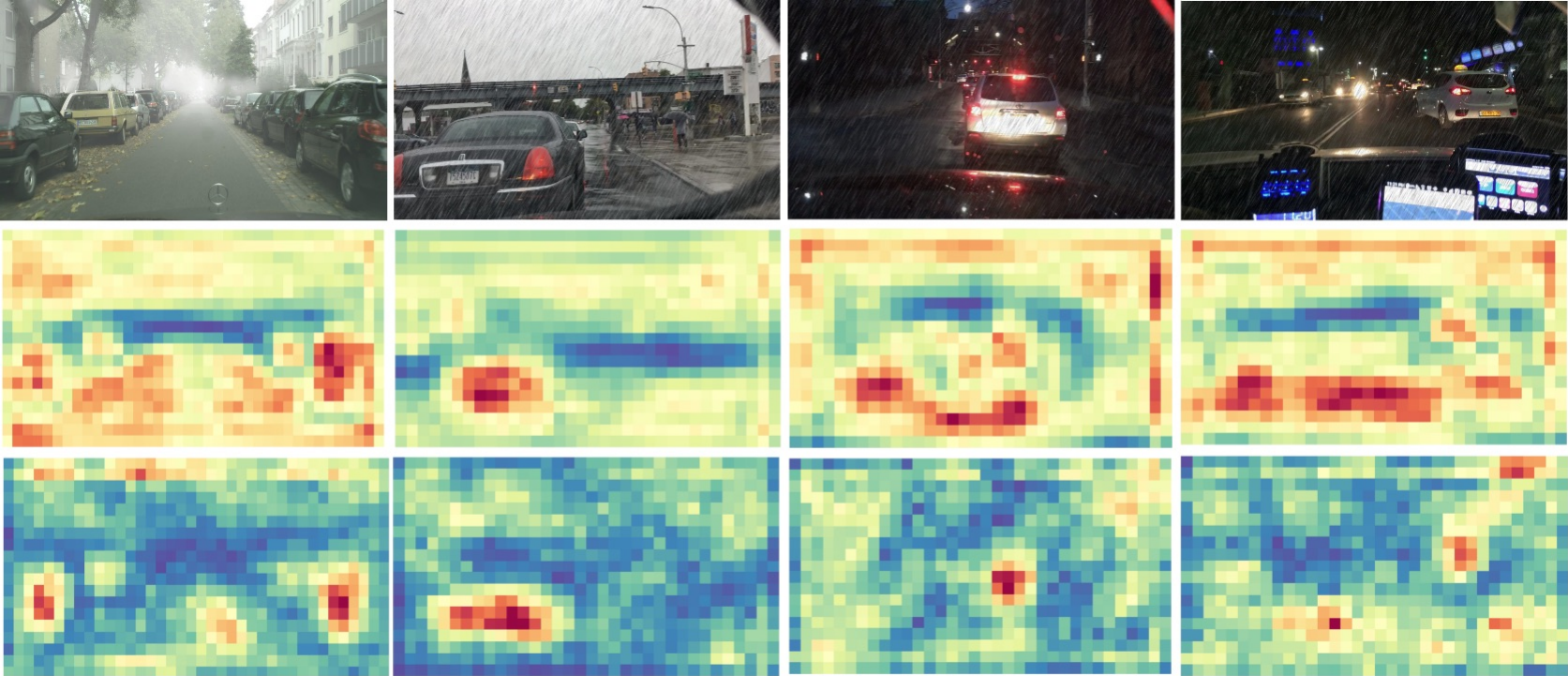}
\end{center}
\vspace{-1.em}
\caption{\textbf{Visualization results of feature maps on Diverse-Weather Dataset~\cite{wu2022single}}. The top row is the input image, the second row is the results of the baseline method SKIP~\cite{yoo2024and}, and the bottom row is the results of our proposed method. 
}
\label{featmap_vis}
\vspace{-1.5em}
\end{figure*}

\textbf{Visualization Analysis.}
We conducted a visualization analysis of the CTTAOD results on the Diverse-Weather dataset, as shown in Fig.~\ref{visual_box}. The top row presents the detection results from the SKIP~\cite{yoo2024and} method and the bottom row presents the results from our proposed method.
It is shown that, compared to the SKIP~\cite{yoo2024and} baseline, our method achieves more accurate classification and localization of objects such as cars, buses, and persons on four challenging street scenarios. The experimental results demonstrate the effectiveness of our parameter-generated tuning for the domain-aware adapter in continual test-time adaptive object detection tasks.

Fig.~\ref{featmap_vis} shows the feature map visualization of layer 4 of the FPN in the backbone on the target domain of Night-Clear. The top row is the input images, the second row is the results of SKIP~\cite{yoo2024and}, and the bottom row is the results of our proposed method. 
The visualization results demonstrate that our proposed domain-aware Adapter with parameter generation effectively focuses attention on the objects, whereas the baseline method, SKIP \cite{yoo2024and}, tends to allocate more attention to areas outside the objects. The feature map visualizations clearly show that the domain-aware adapter can successfully disentangle domain-invariant and domain-specific features, enabling the model to better understand scenes in the target domain.

\vspace{-0.8em}
\subsection{Ablation Study}

\begin{table}[]
\footnotesize
\setlength{\tabcolsep}{0.8pt}
\begin{center}
\begin{tabular}{l|ccc|cc}
\toprule[1pt]
Method  & Adapter & Parameter-Gen & Align &  Diverse-Weather  & COCO  \\ \hline
Direct-Test  &$\usym{2715}$ &$\usym{2715}$ &$\usym{2715}$& 27.9 &17.3  \\ \hline    
Ours  &$\usym{1F5F8}$ & $\usym{2715}$&$\usym{1F5F8}$ & 29.5 & 22.4  \\
Ours  &LoRA \cite{hu2022lora} & $\usym{1F5F8}$ &\usym{1F5F8} & 28.9 & 21.8  \\ 
Ours  &$\usym{1F5F8}$ &$\usym{1F5F8}$& KL & 30.1 & 22.3   \\ 
Ours  &$\usym{1F5F8}$ & $\usym{1F5F8}$& \usym{1F5F8}  & \textbf{30.7} &\textbf{23.0}    \\ 
\bottomrule[1pt]
\end{tabular}
\end{center}
\vspace{-2em}
\caption{Ablation study (\%) on Diverse-Weather dataset with backbone of ResNet-101 and COCO dataset with backbone of ResNet-50. ``KL'' refers to the Kullback-Leibler divergence alignment method.
}
\vspace{-3em}
\label{abation_study}
\end{table}

We conducted a series of ablation studies to analyze the impact of different components of our proposed method. 
Table~\ref{abation_study} shows the results of the ablation experiment. 
It can be seen that when introducing the domain-aware adapter and alignment method for test-time adaptation, the average accuracy of the model reaches 29.5\%, marking a significant improvement over the baseline method.
The average accuracy on the Diverse-Weather dataset and COCO dataset is 28.9\% and 21.8\% when introducing the simple LoRA \cite{hu2022lora} and parameter-generated tuning methods.
The average accuracy on the Diverse-Weather dataset and COCO dataset is 30.1\% and 22.3\% when introducing the domain-aware adapter and the Kullback-Leibler divergence alignment method.
Finally, when introducing the domain-aware adapter with the parameter-generated tuning and class-centered optimal transport alignment method, the generalization performance of our method is further improved, with an average accuracy of 30.7\% and 23.0\% on the Diverse-Weather dataset and COCO dataset, respectively.

\vspace{-1.5em}
\section{Conclusion}
\vspace{-0.5em}
To tackle the key challenges of update efficiency, catastrophic forgetting, and local optima due to continual unsupervised training in the CTTAOD task. In this paper, we proposed a novel parameter-generated tuning method. Specifically, we proposed a dual-path LoRA-based domain-aware adapter to disentangle features into domain-invariant and domain-specific features. Additionally, we propose an environmental conditional diffusion-based parameter-generated tuning method to generate robust parameters with the target scene conditions for the domain-aware adapter. Finally, a class-centered optimal transport alignment is proposed to mitigate catastrophic forgetting. Experimental results on various continual test-time scenarios demonstrate the effectiveness and generalizability of our method.

{
    \small
    \bibliographystyle{ieeenat_fullname}
    \bibliography{main}
}

\end{document}